\documentclass{ieeeaccess}
\usepackage{cite}
\usepackage{amsmath,amssymb,amsfonts}
\usepackage{algorithmic}
\usepackage{graphicx}
\usepackage{textcomp}
\def\BibTeX{{\rm B\kern-.05em{\sc i\kern-.025em b}\kern-.08em
    T\kern-.1667em\lower.7ex\hbox{E}\kern-.125emX}}
\begin{document}

\history{Preprint.}
\doi{10.1109/ACCESS.2017.DOI}

\title{TransMed: Transformers Advance Multi-modal Medical Image Classification}

\author{Yin Dai\authorrefmark{1,2}, Yifan Gao\authorrefmark{1}.} 
\address[1]{College of Medicine and Biological Information Engineering, Northeastern University, China}
\address[2]{Engineering Center on Medical Imaging and Intelligent Analysis, Ministry Education, Northeastern University, Shenyang 110169, China}

\markboth
{Y. Dai \headeretal: TransMed: Transformers Advance Multi-Modal Medical Image Classification}
{Y. Dai \headeretal: TransMed: Transformers Advance Multi-Modal Medical Image Classification}

\corresp{Corresponding author: Yin Dai (e-mail: daiyin@bmie.neu.edu.cn).}

\begin{abstract}
Over the past decade, convolutional neural networks (CNN) have shown very competitive performance in medical image analysis tasks, such as disease classification, tumor segmentation, and lesion detection. CNN has great advantages in extracting local features of images. However, due to the locality of convolution operation, it can not deal with long-range relationships well. Recently, transformers have been applied to computer vision and achieved remarkable success in large-scale datasets. Compared with natural images, multi-modal medical images have explicit and important long-range dependencies, and effective multi-modal fusion strategies can greatly improve the performance of deep models. This prompts us to study transformer-based structures and apply them to multi-modal medical images. Existing transformer-based network architectures require large-scale datasets to achieve better performance. However, medical imaging datasets are relatively small, which makes it difficult to apply pure transformers to medical image analysis. Therefore, we propose TransMed for multi-modal medical image classification. TransMed combines the advantages of CNN and transformer to efficiently extract low-level features of images and establish long-range dependencies between modalities. We evaluated our model for the challenging problem of preoperative diagnosis of parotid gland tumors, and the experimental results show the advantages of our proposed method. We argue that the combination of CNN and transformer has tremendous potential in a large number of medical image analysis tasks. To our best knowledge, this is the first work to apply transformers to medical image classification.
\end{abstract}


\titlepgskip=-15pt

\maketitle

\section{INTRODUCTION}
\label{sec:introduction}
Transformers were first applied in the field of natural language processing (NLP) \cite{b1}. It is a deep neural network mainly based on the self-attention mechanism to extract intrinsic features. Because of its powerful representation capabilities, researchers hope to find a way to apply transformers to computer vision tasks. Compared with text, images involve larger size, noise, and redundant modalities, so it is considered more difficult to use transformers on these tasks. Recently, transformers have made a breakthrough in computer vision. A large number of transformer-based methods have been proposed for computer vision tasks, such as DETR \cite{b2} for object detection, SETR \cite{b3} for semantic segmentation, ViT \cite{b4} and DeiT \cite{b5} for image recognition.

Transformers have achieved success in natural images, but it has received little attention in medical image analysis, especially in multi-modal medical image fusion. Multi-modal images are widely used in medical image analysis to achieve lesion segmentation or disease classification. The existing medical image multi-modal fusion based on deep learning can be divided into three categories: input-level fusion, feature-level fusion, and decision-level fusion \cite{b6}. Input-level fusion strategy fuses multi-modal images into the deep network by multi-channel, learns fusion feature representation, and then trains the network. Input-level fusion can retain the original image information to the maximum extent and learn the image features. Feature-level fusion strategy trains a single deep network by taking the image of each modality as a single input. Each representation is fused in the network layer, and the final result is fed to the decision layer to obtain the final result. Feature-level fusion network can effectively capture the information of different modalities of the same patient. Decision-level fusion integrates the output of each network to obtain the final result. Decision-level fusion network aims to learn more abundant information from different modalities independently.

However, they all have shortcomings in varying degrees. The input-level fusion strategy is difficult to establish the internal relationship between different modalities of the same patient, which leads to the degradation of the model performance. Each modality of the feature-level network corresponds to a neural network, which brings huge computational costs, especially in the case of a large number of modalities. The output of each modality of decision-level fusion is independent of each other, so the model cannot establish the internal relationship between different modalities of the same patient. In addition, like decision-level fusion strategy, decision-level fusion strategy is also computationally intensive.

Therefore, there is an urgent need to combine the three fusion strategies efficiently. A good multi-modal fusion strategy should achieve as much interaction between different modalities as possible with low computational complexity.

Compared with CNN, transformers can effectively mine long-range relationships between sequences. The existing computer vision models based on transformer mainly deal with 2D natural images, such as ImageNet \cite{b6} and other large-scale datasets. The method of constructing sequences in 2D images is to cut the images into a series of patches. This kind of sequence construction method implicitly shows long-range dependencies, which is not very intuitive, so it may be difficult to bring significant performance improvement.

On the contrary, there are more explicit sequences in medical images, which contain important long-range dependency and semantic information, as shown in Fig.\ref{fig1}. Due to the similarity of human organs, most visual representations are orderly in medical images. Destruction of these sequences will significantly reduce the performance of the model. It can be considered that compared with natural images, the sequence relationship of medical images (such as modality, slice, patch) holds more abundant information. In practice, doctors will synthesize the pathological information of each modality to make the diagnosis. However, the existing multi-modal fusion methods are too simple to consider the correlation of these sequences, and lack of modeling for these long-range dependencies. The transformer is an elegant, efficient, and powerful encoder for processing sequence relations, which is the motivation for us to propose the multi-modal medical image classification method based on transformers.

In this work, we present the first study to explore the tremendous potential of transformers in the context of multi-modal medical image classification. The proposed method is inspired by the property that the transformer is effective in extracting the relationship between sequences. However, due to the small scale of medical image datasets and the lack of sufficient information to establish the relationship between low-level semantic features, the performance of pure transformer networks based on ViT and DeiT is not satisfactory in multi-modal medical image classification. Therefore, we propose TransMed, which combines the advantages of CNN and transformer to capture low-level features and cross-modality high-level connections. TransMed first processes the multi-modal images as sequences and sends them to CNN, then uses transformers to learn the relationship between the sequences and make predictions. Since the transformer effectively models the global features of multi-modal images, TransMed outperforms the existing multi-modal fusion methods in terms of parameters, operation speed, and accuracy. A large number of experiments have proved the effectiveness of our method.

\Figure[t!](topskip=0pt, botskip=0pt, midskip=0pt)[width=3in]{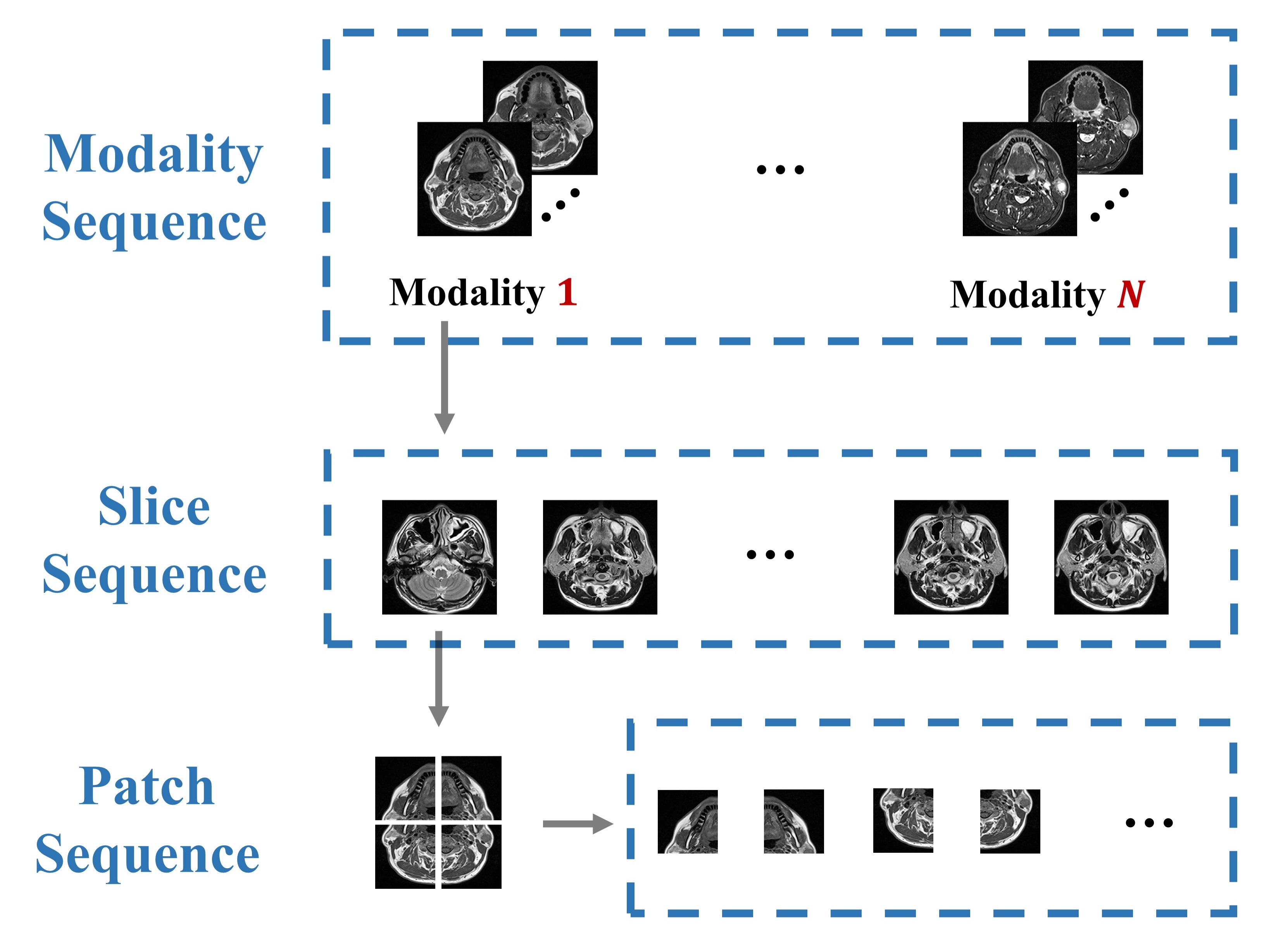}
{Compared with natural images, multi-modal medical images have more informative sequences.\label{fig1}}

In summary, we make the following three contributions:
\begin{enumerate}
	 \item We apply transformers to medical image classification for the first time, and greatly improve the accuracy and efficiency of deep models.
	 \item We propose a novel multi-modal image fusion strategy in this work, which can be leveraged to capture mutual information from images of different modalities in a more efficient way.
     \item Experimental evaluations demonstrate that the proposed method achieves the most advanced performance in the classification of the parotid gland tumors.
\end{enumerate}

The rest of this paper is organized as follows. Section II presents some closely related works. The pipeline of our proposed method is in Section III. Section IV introduces the experimental results and details. Finally, we summarize our work in Section V.

\Figure[t!](topskip=0pt, botskip=0pt, midskip=0pt)[width=6.6in]{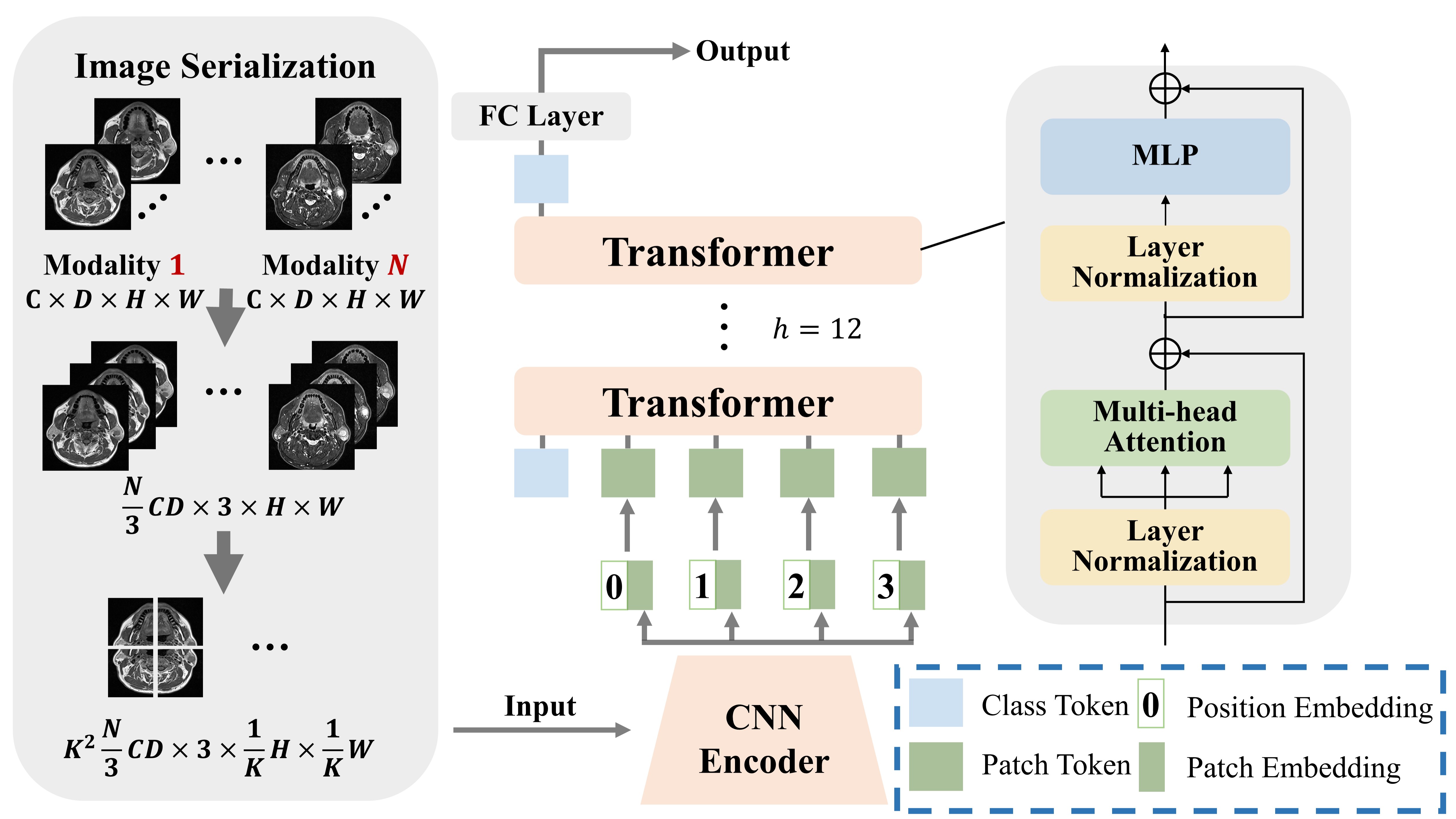}
{Overview of TransMed, which is composed of CNN branch and transformer branch.\label{fig2}}

\section{RELATED WORK}
\subsection{Multi-Modal Medical Image Analysis}
Multi-modal medical analysis is one of the most fundamental and challenging parts of medical image analysis. It is proved that a reasonable fusion of different modalities has been a potential means to enhance Deep networks \cite{b6}. Multi-modal fusion can capture more abundant pathological information and improve the quality of diagnosis.
 
\cite{b8,b9,b10} mainly used the input-level fusion, which is the most common fusion method in multi-modal medical image analysis. Some other papers have shown the potential of feature-level fusion in medical image processing. Hyper DenseNet built dual deep networks for different modalities of Magnetic resonance imaging (MRI) and linked features across these streams \cite{b11}. \cite{b12} fused final features from modality-specific paths to make final decisions. MMFNet used specific encoders to capture modality-specific features and designs a decoder with a complex structure to fuse these features \cite{b13}. Different from the first two techniques, \cite{b14}, \cite{b15} applied decision-level fusion technology to improve performance. \cite{b15} set three modality-specific encoders to capture low-level features and a decoder to fuse low-level and high-level features, then the results of each branch were fused to generate the final result. \cite{b14} designed a gate network to dynamically combine each decision and make a prediction.
 
Besides, some studies have evaluated multiple fusion methods at the same time. \cite{b16} used feature-level fusion and decision-level fusion in their work. \cite{b17} designed three kinds of fusion networks, and gets better performance than a single modality. These fusion methods improve the performance of the model to a certain extent, but there are some shortcomings, such as poor scalability, large computational complexity, and difficulty in establishing long-range connections.
 
\subsection{Transformers}
Transformers were first proposed for machine translation and achieved satisfactory results in a large number of NLP tasks. Then, the transformer structures were introduced into the field of computer vision, and some modifications were made according to the specific tasks. The results show the potential of transformers to surpass pure CNN. Some work uses the framework of CNN and transformer \cite{b2,b18}, while others directly use pure transformers to replace CNN \cite{b2,b4,b5,b19}. These methods have shown encouraging results in computer vision tasks, but their direct applications in multi-modal medical images are not effective and require a lot of computing resources. As far as we know, TransMed is the first multi-modal medical image classification framework based on transformers, which provides a novel multi-modal image fusion strategy.

\section{METHODS}
The most common method of multi-modal medical image classification is to train CNN directly (such as Resnet \cite{b20}). Firstly, the image is encoded as a high-level feature representation, and then its features or decisions are fused. Different from the existing methods, our method uses transformers to introduce the self-attention mechanism into the multi-modal fusion strategy. We will first introduce how to directly apply transformers to aggregate feature representations from decomposed image patches in Section 3.A. Then, the overall framework of TransMed will be described in detail in Section 3.B.

\subsection{Transformers Aggregate Multi-modal Features}
In this work, we follow the original DeiT implementation as much as possible. The advantage of this intentionally simple setting is to reduce the impact of other tricks on the performance of the model and intuitively show the benefits of transformers. In addition, we can use the extensible DeiT model and its pre-trained weights almost immediately.

The important components of the transformer including self-attention (SA), multi-head self-attention (MSA), and multi-layer perception (MLP). The input of transformers includes a variety of embeddings and tokens. Slightly different from DeiT, we remove the linear projection layer and distillation token. We will describe each of these components in this section.

\subsubsection{Self-Attention}

\Figure[t!](topskip=0pt, botskip=0pt, midskip=0pt)[width=1.5in]{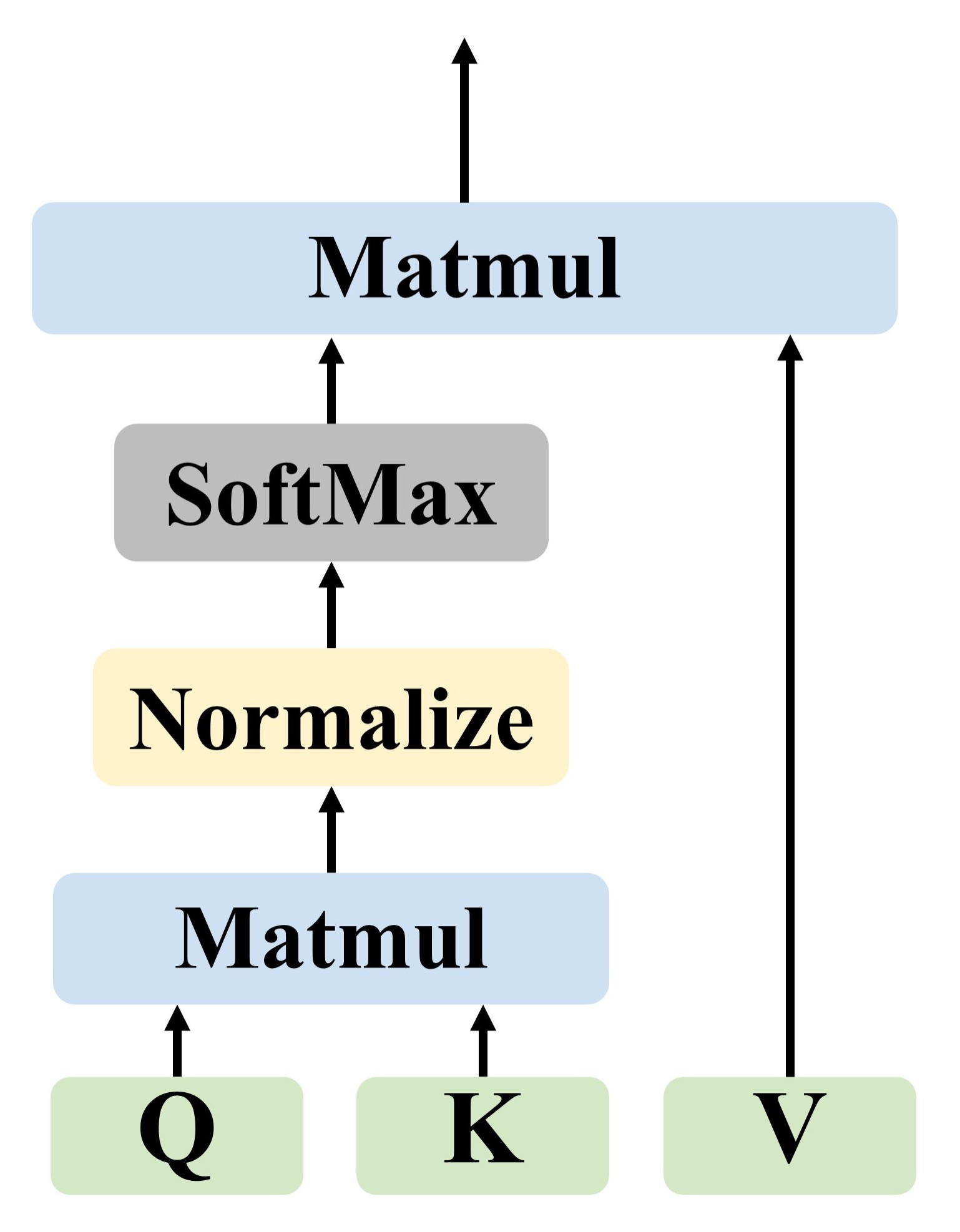}
{Overview of self-attention, matmul means matrix product of two arrays.\label{fig3}}

SA is an attention mechanism, which uses other parts of the same sample to predict the rest of the data sample. In computer vision, it is a little similar to non-local networks \cite{b21}. SA has many forms, and the common transformer relies on the form of scaled dot-product shown in Figure \ref{fig3}. In the SA layer, the input vector is first transformed into three different vectors: query matrix Q, key matrix K, and value matrix V, the output is the weighted sum of the value vectors. The weight assigned to each value is determined by the dot product of the query and the corresponding key. The attention function between different input vectors is calculated as follows:
\begin{equation}
	Attention(Q,K,V)=Softmax({\frac{QK^\mathrm{T}}{\sqrt{d_k}}})\cdot V
\end{equation}

Where $d_k$ is the dimension of key vector $k$. $\sqrt{d_k}$ provides an appropriate normalization to make the gradient more stable.

\subsubsection{Multi-head Self-Attention}
\Figure[t!](topskip=0pt, botskip=0pt, midskip=0pt)[width=3.3in]{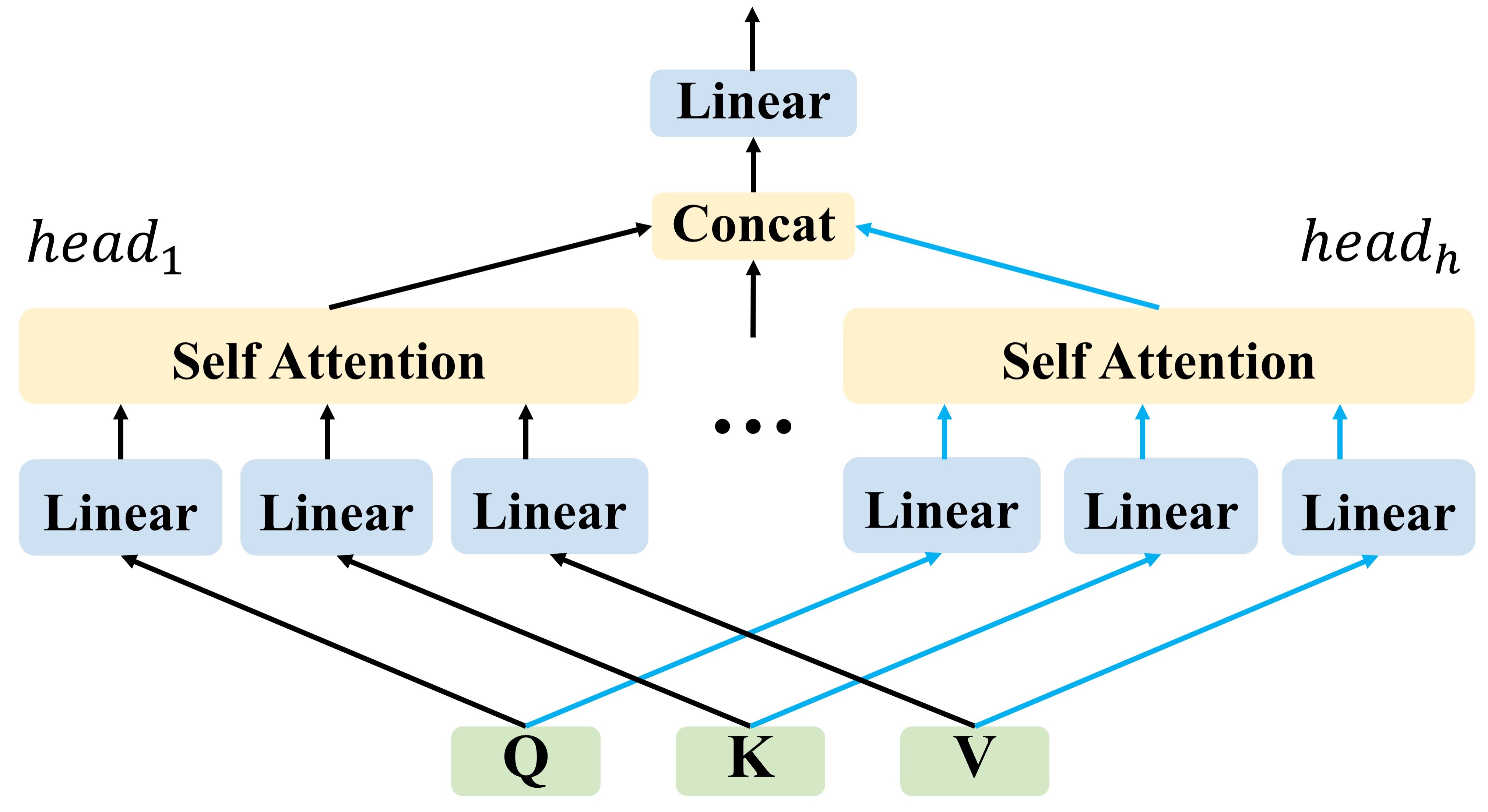}
{An illustration of our multi-head self-attention component, concat means concatenate representations.\label{fig4}}

MSA is the core component of the transformer. As shown in Figure \ref{fig4}, The difference from SA is that the multi-head mechanism splits the input into many small parts, then calculates the scaled dot-product of each input in parallel, and splices all the attention outputs to get the final result. The formula of MSA can be written as follows:
\begin{equation}
	head_i=Attention(QW_i^Q,KW_i^K,VW_i^V)
\end{equation}

\begin{equation}
	MSA(Q,K,V)=Concat(head_i,...,head_i)W^O
\end{equation}

Where the projections $W_i^Q$, $W_i^K$, $W_i^V$ and $W^O$ are trainable parameter matrices, $h$ is the number of transformer layers. The advantage of MSA is that it allows the model to learn sequence and location information in different representation subspaces.

\subsubsection{Multi-Layer Perceptron}
In this paper, a MLP is added on top of the MSA layer. The MLP is composed of linear layers separated by a GeLU \cite{b22} activation. Both MSA and MLP have skip-connections like residual networks and with a layer normalization. Therefore, it is assumed that the representation of the $t-1$ layer is $x_{t-1}$, LN represents the linear normalization, and the output of the $t$ layer can be written as follows:

\begin{equation}
	\hat{x_t{}}=MSA(LN(x_{t-1}))+x_{t-1}
\end{equation}

\begin{equation}
	x_t=MLP(LN(\hat{x_t{}}))+\hat{x_t{}}
\end{equation}

\subsubsection{Embeddings and Tokens}
The input layer contains five embeddings and tokens, which are patch embedding, position embedding, class embedding, patch token, and class token.

Patch embedding is the representation of each patch's output from CNN, and class embedding is a trainable vector. To encode the spatial information and location information of a patch into patch tokens, we use position embeddings and patch embeddings to preserve the information. Class embedding does not have patch embedding that can be added, so class token and class embedding are equivalent. Suppose the input is $x$, the trainable vector is $W^c$, the position embedding is $x_{po}$, patch tokens $x_{pt}$ and class token $x_{ct}$ can be expressed as follows:

\begin{equation}
	x_{pt}=Conv(x)+x_{po}
\end{equation}

\begin{equation}
	x_{ct}=W^c
\end{equation}

The class token is attached to patch tokens before the input layer of transformers, passes through the transformer layer, and then outputs from the fully connected layer to predict the class.

\subsection{TransMed}
The structure of TransMed is shown in Figure \ref{fig2}. Instead of using pure transformers as the encoder, TransMed adopts a hybrid model including CNN and transformer, in which CNN is used as a low-level feature extractor to generate the patch embedding.

Given a multi-modal image $x \in R^{B \times M \times C \times D \times H \times W}$,  where spatial resolution is $H \times W$, the depth is $D$, the number of channels is $C$, the number of modalities is $M$, and the batch size is $B$. Before sending it to the CNN encoder, it is necessary to construct the sequence. Firstly, three adjacent 2D slices of a multi-modal image are superimposed to construct three-channel images. Then, according to \cite{b4}, each image will be divided into $K \times K$. The larger $K$ value means that the size of each patch is smaller. We will evaluate the impact of different $K$ values on the performance of the model in Section 4.E. Finally, the image is encoded into a  $(\frac{1}{3}BMCDK^2, 3, \frac{H}{K}, \frac{W}{K})$ patch.

After the image sequence is constructed, it is input into the 2D CNN. The last fully connected layer of 2D CNN is replaced by a linear projection layer to map the features of the vector patch to the potential embedding space. 2D CNN extracts low-level features from the image sequence and encodes them preliminarily. The output shape is $(B,\frac{1}{3}MC DK^2,P)$, in which the size of $P$ is set to adapt to the input size of the transformer.

\begin{table*}
	
	\caption{Comparison on the parotid gland tumors dataset (average accuracy $\%$ and precision $\%$ for each disease. IF, FF and DF represent input-level fusion, feature-level fusion and decision-level fusion, respectively.).
	}
	\label{table1}

	\renewcommand{\arraystretch}{1.5}
	\begin{tabular}{cccccccccccc}
		\hline
		\textbf{Method}&
		\textbf{Dim}&
		\textbf{Backbone}&
		\textbf{Fusion}&
		\textbf{Params}&
		\textbf{TFlops}&
		\textbf{Acc}&
		\textbf{PA}&
		\textbf{WT}&
		\textbf{MT}&
		\textbf{BCA}&
		\textbf{OBL}
		\\
		\hline
		
		P3D         & 3D & P3D              & IF & 67M  & 0.31 & 76.1±5.5 & 59.9±23.1 & 84.3±5.3  & 69.7±19.0 & 71.4±7.3  & 78.0±14.0 \\
		C3D         & 3D & ConvNet          & IF & 28M  & 1.36 & 71.0±4.1 & 68.3±38.9 & 81.3±15.4 & 67.8±10.0 & 71.4±7.5  & 84.5±12.4 \\
		Resnet34    & 2D & Resnet34         & IF & 22M  & 0.02 & 69.9±4.0 & 81.0±11.0 & 77.6±7.4  & 61.4±10.8 & 53.8±15.5 & 68.1±7.9  \\
		Resnet152   & 2D & Resnet152        & IF & 58M  & 0.05 & 69.0±3.5 & 50.5±18.0 & 74.1±10.1 & 64.3±20.0 & 62.9±8.9  & 75.2±11.4 \\
		3D Resnet34 & 3D & 3D Resnet34      & IF & 64M  & 1.08 & 73.3±5.1 & 69.2±16.2 & 86.8±5.2  & 75.3±18.1 & 68.7±13.8 & 65.7±6.4  \\
		\cite{b11}    & 3D & ConvNet          & FF & 45M  & 3.47 & 74.2±2.9 & 76.0±24.8 & 86.2±10.0 & 72.9±16.3 & 75.2±24.6 & 80.0±15.2 \\
		\cite{b12}    & 3D & 3D Resnet34      & FF & 130M & 1.35 & 73.3±2.4 & 46.2±13.2 & 78.4±7.9  & 70.2±15.5 & 69.8±15.8 & 79.0±10.7 \\
		3D Resnet34 & 3D & 3D Resnet34      & DF & 128M & 1.28 & 72.1±3.5 & 64.7±14.4 & 81.5±9.3  & 66.8±8.4  & 69.6±8.9  & 72.1±14.9 \\
		P3D         & 3D & P3D              & DF & 136M & 0.42 & 74.8±4.6 & 50.5±20.0 & 85.1±4.4  & 70.5±20.2 & 69.5±8.7  & 73.4±14.3 \\
		C3D         & 3D & ConvNet          & DF & 57M  & 1.45 & 71.0±3.3 & 58.3±33.3 & 70.7±9.3  & 74.0±8.5  & 78.9±20.0 & 73.2±6.7  \\
		Resnet34    & 2D & Resnet34         & DF & 45M  & 0.03 & 71.3±4.5 & 72.7±21.7 & 75.3±8.1  & 72.5±10.3 & 60.9±16.8 & 70.3±9.7  \\
		Resnet152   & 2D & Resnet152        & DF & 116M & 0.06 & 72.2±5.5 & 63.5±18.3 & 75.6±10.4 & 73.4±18.7 & 83.2±16.3 & 69.6±11.5 \\
		\hline
		TransMed-T  & 2D & Renset18+DeiT-T  & —— & 17M  & 0.09 & 87.0±2.6 & 80.1±13.8 & 87.3±3.0  & 90.7±5.1  & 82.5±15.3 & \textbf{93.6±3.3}  \\
		\textbf{TransMed-S}  & 2D & Resnet34+DeiT-S  & —— & 43M  & 0.19 & \textbf{88.9±3.0} & \textbf{90.1±12.2} & \textbf{89.2±6.8}  & \textbf{92.0±4.4}  & 82.9±9.3  & 88.3±6.1  \\
		TransMed-B  & 2D & Resnet50+DeiT-B  & —— & 110M & 0.22 & 87.4±2.1 & 86.2±15.2 & 88.4±3.8  & 88.2±7.0  & \textbf{84.8±13.8} & 92.2±8.0  \\
		TransMed-L  & 2D & Resnet152+DeiT-B & —— & 145M & 0.58 & 86.6±3.4 & 85.4±12.9 & 88.7±6.1  & 90.4±5.3  & 75.9±8.0  & 91.8±7.5 
		\\
		
		\hline
		
	\end{tabular}
	\label{tab1}
\end{table*}

\section{RESULTS}
\label{sec:guidelines}

\subsection{DATASET AND PREPROCESSING}
\subsubsection{DATASET}
We use a dataset collected in cooperative hospitals to evaluate the performance of our proposed method under multi-modal images. This dataset included two modalities of MRI (T1 and T2) of 344 patients, and the ground truth labels are obtained from biopsies.

The incidence of malignant tumors in parotid gland tumors is about 20$\%$ \cite{b23}. Correct preoperative diagnosis of these tumors is essential for proper surgical planning. Among them, imaging examination plays an important role in determining the nature of parotid gland masses. MRI is considered to be the preferred imaging method for preoperative diagnosis of parotid tumors \cite{b24}. MRI can provide information about the exact location of the lesion, the relationship with the surrounding structure, and can assess the spread of nerves and bone invasion. However, it is reported that parotid gland tumors show considerable overlap in imaging features (such as tumor margins, homogeneity, and signal intensity), so it is difficult for doctors to identify the mass.

According to common clinical classifications, we divide parotid gland tumors into five categories: Pleomorphic Adenoma (PA), Warthin Tumor (WT), Malignant Tumor (MT), Basal Cell Adenoma (BCA), and Other Benign Lesions (OBL) \cite{b25}.

\subsubsection{PREPROCESSING}
First, perform OTSU [26] to extract the foreground area in the original image. Then the images of different modalities of the same patient are registered to improve the consistency of the foreground area. Then resample each image to (18, 448, 448). Therefore, 344 images are finally included, each of which is a stack of 3D images of MRI T1 and T2, and the size is (36, 448, 448). Data augmentation uses random flipping and random noise. Random flipping performs flipping of the image with 50$\%$ probability. Random noise adds Gaussian noise with a mean value of 0 and a variance of 0.1 to the image.

\subsection{Experimental Settings and Evaluation Criteria}
The patients were randomly divided into the training group (n = 275) and independent test group (n = 69) according to the ratio of 4:1, and then the training group was used to optimize the model parameters. We set SGD as the optimizer with a learning rate equal to  $10^{-3}$ and momentum equal to 0.7. The maximum training round is set to 100. Our experiments were performed on NVIDIA 3080 GPU (with 10GB GPU memory). The code is implemented using Pytorch \cite{b27} and TorchIO \cite{b28}. To eliminate accidental factors, each model is subjected to 10 independent experiments, and each experiment is randomly divided into the training group and the test group. Besides, other experimental parameters keep consistent during training.

The evaluation criteria for each model are the overall accuracy rate $ACC(i)$ and the precision rate of each category $P(i)$, as defined in the following:

\begin{equation}
	ACC=\frac{T_c}{T}
\end{equation}

Where $T$ is the total number of samples and $T_c$ is the total number of samples with the correct prediction.

\begin{equation}
	P(i)=\frac{T_{ic}}{T_{ic}+T_{if}}
\end{equation}

Where $T_{ic}$ is the total number of samples correctly predicted as class $i$, and $T_{if}$ is the total number of samples that are wrongly predicted as class $i$. $P(i)$ can describe the stability and robustness of the model in small datasets.

\subsection{Baseline Methods}
The input-level fusion strategy can be easily implemented using mainstream 2D CNN and 3D CNN, so the selected network includes Resnet34, Resnet152, 3D Resnet34, P3D \cite{b29}, and C3D \cite{b30}. In feature-level fusion experiments, we used two common feature-level fusion methods \cite{b11,b12}. Since these two papers focus on segmentation tasks, we modify the network structure to adapt to the classification tasks. The deep network used in the decision-level fusion experiments is the same as the input-level strategy.

\subsection{Results}
Table. 1 reports the performance of our proposed models, in which four variants are provided: the tiny version (TransMed-T) use ResNet18 and DeiT-Tiny (DeiT-T) as backbones for CNN branch and transformer branch, respectively; the small version (TransMed-S) use ResNet34 and DeiT-Small (DeiT-S) as backbone; the base version (TransMed-B) use ResNet50 and DeiT-Base (DeiT-B) as backbone; the large version (TransMed-L) use ResNet152 and DeiT-B.

TransMed consistently outperforms previous multi-modal fusion strategies by a large margin. TransFuse-S achieves on average about 12.8$\%$ improvement in terms of the average accuracy with respect to the P3D while the larger version TransMed-B and TransMed-L slightly suffer from overfitting on the dataset. Table \ref{tab1} also compares the number of parameters and computational costs between our proposed models and previous methods. TransMed achieves state-of-the-art performance with much fewer parameters and computational costs. TransMed is highly efficient as it models the long-range relationship between modalities very well. We expect that our method can inspire further exploration of multi-modal medical image fusion in future work.

\begin{table*}
	\caption{Ablation study on the effectiveness of CNN branch and transformer branch.
	}
	\label{table2}
	\renewcommand{\arraystretch}{1.5}
	\begin{tabular}{ccccccccc}
		\hline
		Model           & Params & TFlops & Acc      & PA        & WT        & MT        & BCA       & OBL       \\
		\hline
		TransMed-T      & 17M    & 0.09   & 87.0±2.6 & 80.1±13.8 & 87.3±3.0  & 90.7±5.1  & 82.5±15.3 & 93.6±3.3  \\
		w/o transformer & 12M    & 0.01   & 71.3±2.5 & 71.6±17.8 & 78.9±3.9  & 73.9±12.5 & 66.9±13.1 & 69.9±13.3 \\
		w/o CNN         & 5M     & 0.07   & 51.3±5.9 & 20.0±18.7 & 61.5±13.5 & 37.0±5.7  & 41.7±40.1 & 51.0±16.4 \\
		\hline
	\end{tabular}
	\label{tab2}
\end{table*}

\begin{table*}
	\caption{Ablation study on different patch sizes.
	}
	\label{table3}
	\renewcommand{\arraystretch}{1.5}
	\begin{tabular}{cccccccc}
		\hline
		Model      & K  & Acc      & PA        & WT       & MT        & BCA       & OBL       \\ \hline
		TransMed-T & 1  & 86.8±2.3 & 83.1±12.4 & 90.1±3.9 & 88.7±10.1 & 78.3±13.7 & 95.3±5.9  \\
		TransMed-T & 2  & 87.0±2.6 & 80.1±13.8 & 87.3±3.0 & 90.7±5.1  & 82.5±15.3 & 93.6±3.3  \\
		TransMed-T & 4  & 86.4±3.3 & 86.3±17.2 & 87.0±6.5 & 89.5±3.6  & 75.5±8.3  & 92.1±7.2  \\
		TransMed-T & 8  & 80.0±5.8 & 81.2±17.4 & 81.1±5.3 & 88.5±4.2  & 63.9±12.1 & 88.2±13.6 \\
		TransMed-T & 16 & 65.2±5.8 & 49.0±32.2 & 72.1±7.7 & 70.4±7.9  & 62.3±23.0 & 64.1±14.5 \\ \hline
	\end{tabular}
	\label{tab3}
\end{table*}

\subsection{Ablation Experiments}
To demonstrate the effect of transformers in TransMed, we conducted ablation experiments. For TransMed, changing the backbone from TransMed-T to TransMed-S results in 1.9$\%$ improvement in average accuracy, at the expense of a much larger computational cost. Therefore, considering the computation cost, all experimental comparisons in this paper are conducted with TransMed-T to demonstrate the effectiveness of TransMed.

In the experiment, TransMed's CNN and transformers were removed respectively, and all other conditions remained unchanged. The results are shown in Table \ref{tab2}. The results indicate that the transformer greatly improves the ability of the deep model to explore the relationship between modalities with little increase of parameters and computation. However, the performance of the pure transformer structure is poor due to the small dataset.

We also explored the impact of different patch sizes on performance in image serialization by changing $K$ values respectively while other conditions remain unchanged. The results are shown in Table \ref{tab3}. The experimental results show that the performance is poor when the $K$ value is large. The possible reason is that too small image patches destroy the semantic information of the image.

\section{CONCLUSION}
The transformer is a powerful deep neural network structure for processing sequences in NLP, but it has received little attention in medical image analysis. In this paper, we propose TransMed, which is a novel design of multi-modal medical image classification based on transformers. TransMed has achieved very competitive results in challenging parotid tumor classification. TransMed is easy to implement and has a flexible structure, which can be extended to multiple medical image modalities with low resource cost.

These preliminary results are encouraging, but there are still many challenges. One is to apply TransMed to other medical image analysis tasks, such as tumor segmentation and lesion detection. Another challenge is to use the pure transformer structure. Pure transformer structure has been successful in large-scale natural image datasets. However, our preliminary experiments show that there is still a big gap between the pure transformer and typical CNN in small medical image datasets. We expect future work to further improve TransMed.

\vspace{-60 mm}

\EOD

\end{document}